\documentclass[10pt,twocolumn,letterpaper]{article}

\usepackage[pagenumbers]{cvpr} % To force page numbers, e.g. for an arXiv version

\usepackage[dvipsnames]{xcolor}
\usepackage{colortbl}

\definecolor{placeholder}{rgb}{0.6,0.8,0.95}

\newcommand{\modelname}{\emph{Scaffold}-GS\xspace}

\definecolor{amber}{rgb}{1.0, 0.75, 0.0}

\definecolor{visible-blue}{rgb}{0.286, 0.525, 0.910}

\definecolor{tabfirst}{rgb}{1, 0.7, 0.7} % red
\definecolor{tabsecond}{rgb}{1, 0.85, 0.7} % orange
\definecolor{tabthird}{rgb}{1, 1, 0.7} % yellow
\usepackage{overpic}
\usepackage{wrapfig}
\usepackage{amsmath}
\usepackage{empheq}
\usepackage{comment}
\usepackage{makecell}
\usepackage{multirow}
\usepackage{multicol}
\usepackage{float}
\usepackage{marvosym}

\definecolor{cvprblue}{rgb}{0.21,0.49,0.74}
\usepackage[pagebackref,breaklinks,colorlinks,citecolor=cvprblue]{hyperref}

\def\paperID{****} % *** Enter the Paper ID here
\def\confName{CVPR}
\def\confYear{2024}

\title{\modelname: Structured 3D Gaussians for View-Adaptive Rendering}

\author{Tao Lu $^{1,3}$\thanks{*Equal contribution} \quad Mulin Yu$^{1}$\protect\footnotemark[1] \quad Linning Xu$^2$ \quad Yuanbo Xiangli$^4$ \\
Limin Wang$^{1,3}$ \quad Dahua Lin$^{1,2}$ \quad Bo Dai$^{1}$\\
$^1$Shanghai Artificial Intelligence Laboratory, $^2$The Chinese University of Hong Kong, \\
$^3$Nanjing University, $^4$Cornell University
}
\begin{document}

\twocolumn[{%
	\renewcommand
	\twocolumn[1][]{#1}%
	\maketitle
	\begin{center}
		\centering
		\vspace{-15pt}
            \includegraphics[width=\textwidth]{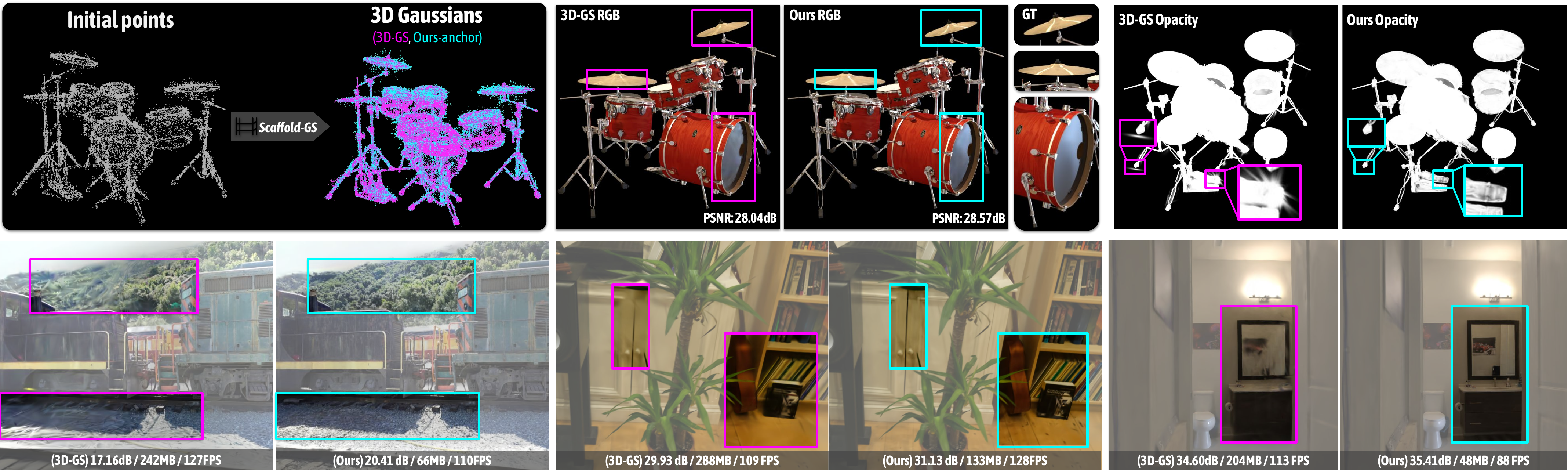}
		\captionof{figure}{\small
            \textbf{\modelname} represents the scene using a set of 3D Gaussians structured in a dual-layered hierarchy. Anchored on a sparse grid of initial points, a modest set of neural Gaussians are spawned from each anchor to \emph{dynamically adapt to} various viewing angles and distances. Our method achieves rendering quality and speed comparable to 3D-GS with a more compact model (last row metrics: PSNR/storage size/FPS). Across multiple datasets, \modelname demonstrates more robustness in large outdoor scenes and intricate indoor environments with challenging observing views~\eg transparency, specularity, reflection, texture-less regions and fine-scale details.
		}
		\label{fig:teaser}
	\end{center}
}]
\maketitle

\begingroup
\renewcommand\thefootnote{}\footnotetext{* denotes equal contribution.}
\endgroup
\begin{abstract}
% Photo-realistic and real-time rendering of 3D scenes is essential in both academia and industry. 
% Traditional primitive-based representations are superior in rendering speed but often yield low-quality renderings with discontinuity and visual artifacts. Volumetric representations use learning-based parametric models, and produce detailed and continuous rendering results, but at the expense of time-consuming stochastic sampling.
\vspace{-5mm}
Neural rendering methods have significantly advanced photo-realistic 3D scene rendering in various academic and industrial applications. The recent 3D Gaussian Splatting method has achieved the state-of-the-art rendering quality and speed combining the benefits of both primitive-based representations and volumetric representations. 
However, it often leads to heavily redundant Gaussians that try to fit every training view, neglecting the underlying scene geometry. Consequently, the resulting model becomes less robust to significant view changes, texture-less area and lighting effects.
We introduce Scaffold-GS, which uses anchor points to distribute local 3D Gaussians, and predicts their attributes on-the-fly based on viewing direction and distance within the view frustum. 
Anchor growing and pruning strategies are developed based on the importance of neural Gaussians to reliably improve the scene coverage. We show that our method effectively reduces redundant Gaussians while delivering high-quality rendering. 
We also demonstrates an enhanced capability to accommodate scenes with varying levels-of-detail and view-dependent observations, without sacrificing the rendering speed. Project page: \href{https://city-super.github.io/scaffold-gs/}{https://city-super.github.io/scaffold-gs/}.
\end{abstract}    
\section{Introduction}
\label{sec:intro}

% Problems of explicit, implicit and 3DGS
Photo-realistic and real-time rendering of 3D scenes has always been a pivotal interest in both academic research and industrial domains, with applications spanning virtual reality \cite{VRNeRF}, media generation \cite{DBLP:conf/iclr/PooleJBM23},  and large-scale scene visualization \citep{Turki_2022_CVPR,tancik2022block,xiangli2022bungeenerf}. 
Traditional primitive-based representations like meshes and points \citep{lassner2021pulsar,Munkberg_2022_CVPR,botsch2005high,yifan2019differentiable} are faster due to the use of rasterization techniques tailored for modern GPUs. 
However, they often yield low-quality renderings, exhibiting discontinuity and blurry artifacts. 
In contrast, volumetric representations and neural radiance fields utilize learning-based parametric models \citep{mildenhall2021nerf,Barron_2021_ICCV,Barron_2023_ICCV},
hence can produce continuous rendering results with more details preserved. 
Nevertheless, they come with the cost of time-consuming stochastic sampling, leading to slower performance and potential noise.

In recent times, 3D Gaussian Splatting (3D-GS) \cite{kerbl3Dgaussians} has achieved state-of-the-art rendering quality and speed. Initialized from point clouds derived from Structure from Motion (SfM)~\citep{sfm}, this method optimizes a set of 3D Gaussians to represent the scene. 
It preserves the inherent continuity found in volumetric representations, whilst facilitating rapid rasterization by splatting 3D Gaussians onto 2D image planes. 

While this approach offers several advantages, it tends to excessively expand Gaussian balls to accommodate every training view, thereby neglecting scene structure. This results in significant redundancy and limits its scalability, particularly in the context of complex large-scale scenes.
Furthermore, view-dependent effects are baked into individual Gaussian parameters with little interpolation capabilities, making it less robust to substantial view changes and lighting effects.

We present \modelname, a Gaussian-based approach that utilizes anchor points to establish a hierarchical and region-aware 3D scene representation. We construct a sparse grid of anchor points initiated from SfM points.
Each of these anchors tethers a set of neural Gaussians with learnable offsets, whose attributes (\ie opacity, color, rotation, scale) are dynamically predicted based on the anchor feature and the viewing position.
Unlike the vanilla 3D-GS which allows 3D Gaussians to freely drift and split,
our strategy exploits scene structure to guide and constrain the distribution of 3D Gaussians, whilst allowing them to \emph{locally} adaptive to varying viewing angles and distances. We further develop the corresponding growing and pruning operations for anchors to enhance the scene coverage.

Through extensive experiments, we show that our method delivers rendering quality on par with or even surpassing the original 3D-GS. At inference time, we limit the prediction of neural Gaussians to anchors within the view frustum, and filter out trivial neural Gaussians based on their opacity with a filtering step~(\ie learnable selector).  
As a result, our approach can render at a similar speed (around $100$ FPS at 1K resolution) as the original 3D-GS with little computational overhead. Moreover, our storage requirements are significantly reduced as we only need to store anchor points and MLP predictors for each scene.

In conclusion, our contributions are: 1) Leveraging scene structure, we initiate \emph{anchor points} from a sparse voxel grid to guide the distribution of local 3D Gaussians, forming a hierarchical and region-aware scene representation; 2) Within the view frustum, we predict \emph{neural} Gaussians from each anchor \emph{on-the-fly} to accommodate diverse viewing directions and distances, resulting in more robust novel view synthesis; 3) We develop a \emph{more reliable} anchor growing and pruning strategy utilizing the predicted neural Gaussians for better scene coverage.
\section{Related work}
\label{sec:Related_work}

\paragraph{MLP-based Neural Fields and Rendering.} Early neural fields typically adopt a multi-layer perceptron (MLP) as the global approximator of 3D scene geometry and appearance. 
They directly use spatial coordinates (and viewing direction) as input to the MLP and predict point-wise attribute,~\eg signed distance to scene surface (SDF)~\citep{Park2019DeepSDFLC,wang2021neus,yariv2021volume,Oechsle2021UNISURFUN}, or density and color of that point~\citep{mildenhall2021nerf,Barron2021MipNeRFAM,xiangli2022bungeenerf}. 
Because of its volumetric nature and inductive bias of MLPs, this stream of methods achieves the SOTA performance in novel view synthesis.
The major challenge of this scene representation is that the MLP need to be evaluated on a large number of sampled points along each camera ray. Consequently, rendering becomes extremely slow, with limited scalability towards complex and large-scale scenes. Despite several works have been proposed to accelerate or mitigate the intensive volumetric ray-marching,~\eg using proposal network~\citep{barron2022mipnerf360}, baking technique~\citep{Hedman2021BakingNR,chen2022mobilenerf}, and surface rendering~\cite{Sitzmann2021LightFN}. They either incorporated more MLPs or traded rendering quality for speed.

\paragraph{Grid-based Neural Fields and Rendering.} This type of scene representations are usually based on a dense uniform grid of voxels. They have been greatly used in 3D shape and geometry modeling~\citep{Choy20163DR2N2AU,Tulsiani2017MultiviewSF,Kar2017LearningAM, Peng2020ConvolutionalON,Genova2019LocalDI,zhao2019localization,Mescheder2018OccupancyNL}.
Some recent methods have also focused on faster training and inference of radiance field by exploiting spatial data structure to store scene features, which were interpolated and queried by sampled points during ray-marching. For instance, Plenoxel~\citep{yu2022plenoxels} adopted a sparse voxel grid to interpolate a continuous density field, and represented view-dependent visual effects with Spherical Harmonics. 
The idea of tensor factorization has been studied in multiple works~\citep{Chen2022TensoRFTR,xu2023grid,Xiangli2023AssetFieldAM,Chen2023FactorFA} to further reduce data redundancy and speed-up rendering. 
K-planes~\citep{FridovichKeil2023KPlanesER} used neural planes to parameterize a 3D scene, optionally with an additional temporal plane to accommodate dynamics. 
Several generative works~\citep{Chan2021EfficientG3,Shue20223DNF} also capitalized on triplane structure to model a 3D latent space for better geometry consistency. InstantNGP~\citep{muller2022instant} used a hash grid and achieved drastically faster feature query, enabling real-time rendering of neural radiance field. Although these approaches can produce high-quality results and are more efficient than global MLP representation, they still need to query many samples to render a pixel, and struggle to represent empty space effectively.

\paragraph{Point-based Neural Fields and Rendering.} Point-based representations utilize the geometric primitive~(\ie point clouds) for scene rendering. A typical procedure is to rasterize an unstructured set of points using a fixed size, and exploits specialized modules on GPU and graphics APIs for rendering~\citep{Botsch2005HighqualitySS,sainz2004point,Ren2002ObjectSE}. In spite of its fast speed and flexibility to solve topological changes, they usually suffer from holes and outliers that lead to artifacts in rendering.  
To alleviate the discontinuity issue, differentiable point-based rendering has been extensively studied to model objects geometry~\citep{gross2011point,insafutdinov18pointclouds,lin2018learning,yifan2019differentiable,Wiles2019SynSinEV}. In particular,~\citep{yifan2019differentiable,Wiles2019SynSinEV} used differentiable surface splatting that treats point primitives as discs, ellipsoids or surfels that are larger than a pixel.~\citep{Aliev2019NeuralPG,Kopanas2021PointBasedNR} augmented points with neural features and rendered using 2D CNNs. As a comparison, Point-NeRF~\citep{xu2022point} achieved high-quality novel view synthesis utilizing 3D volume rendering, along with region growing and point pruning during optimization. However, they resorted to volumetric ray-marching, hence hindered display rate. 
Notably, the recent work 3D-GS~\citep{kerbl3Dgaussians} employed anisotropic 3D Gaussians initialized from structure from motion (SfM) to represent 3D scenes, where a 3D Gaussian was optimized as a volume and projected to 2D to be rasterized as a primitive. Since it integrated pixel color using $\alpha$-blender, 3D-GS produced high-quality results with fine-scale detail, and rendered at real-time frame rate.
\section{Methods} 
\label{sec:methods}

\begin{figure*}[t]
  \centering
   \includegraphics[width=1\linewidth]{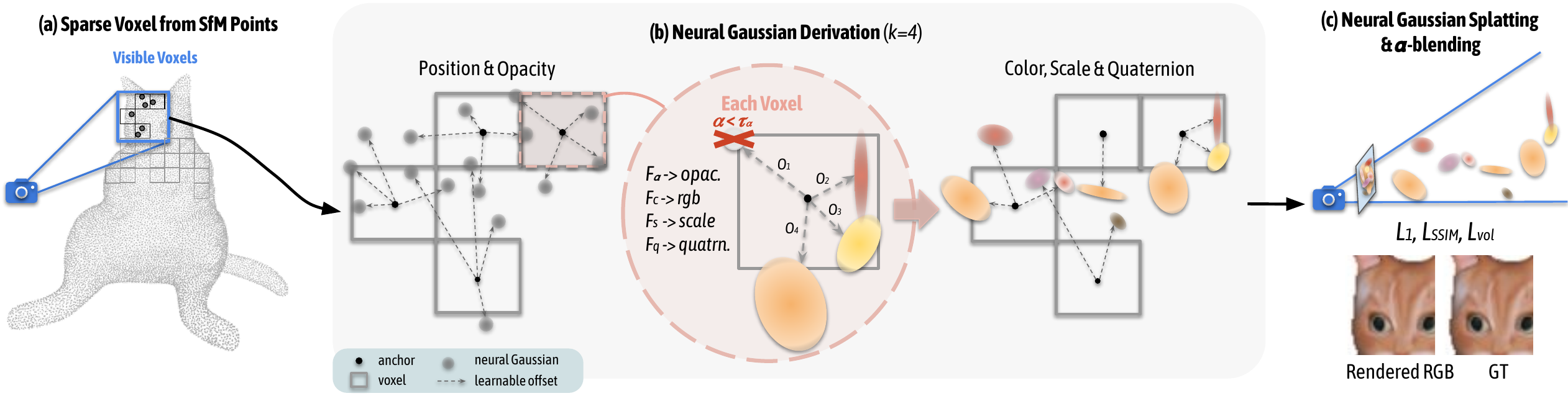}
   \caption{\textbf{Overview of \modelname}. (a) We start by forming a \emph{sparse voxel grid} from SfM-derived points. An \textbf{anchor} associated with a learnable scale is placed at the center of each voxel, roughly sculpturing the scene occupancy. (b) Within a view frustum, \textbf{$k$ neural Gaussians} are spawned from each \emph{visible anchor} with offsets $\{\mathcal{O}_k\}$. Their attributes,~\ie opacity, color, scale and quaternion are then decoded from the anchor feature, relative camera-anchor viewing direction and distance using $F_{\alpha}, F_c, F_s, F_q$ respectively.  
   (c) Note that to alleviate redundancy and improve efficiency, only non-trivial neural Gussians~(\ie $\alpha \geq \tau_\alpha$) are rasterized following~\citep{kerbl3Dgaussians}. The rendered image is supervised via reconstruction ($\mathcal{L}_1$), structural similarity ($\mathcal{L}_{SSIM}$) and a volume regularization ($\mathcal{L}_{vol}$). 
}

   \label{fig:pipeline}
\end{figure*}

The original 3D-GS~\citep{kerbl3Dgaussians} optimizes Gaussians to reconstruct every training view, with heuristic splitting and pruning operations but in general neglects the underlying scene structure. This often leads to highly redundant Gaussians and makes the model less robust to novel viewing angles and distances. To address this issue,
we propose a hierarchical 3D Gaussian scene representation that respects the scene geometric structure, with \emph{anchor points} initialized from SfM to encode local scene information and spawn local \emph{neural Gaussians}. The physical properties of neural Gaussians are decoded from the learned anchor features in a view-dependent manner \emph{on-the-fly}.
Fig.~\ref{fig:pipeline} illustrates our framework. 
We start with a brief background of 3D-GS then unfold our proposed method in details. Sec.~\ref{sec:Anchor} introduces how to initialize the scene with a regular sparse grid of anchor points from the irregular SfM point clouds. Sec.~\ref{sec:Neural} explains how we predict neural Gaussians properties based on anchor points and view-dependent information. To make our method more robust to the noisy initialization, Sec.~\ref{sec:intern} introduces a neural Gaussian based ``growing" and ``pruning" operations to refine the anchor points. Sec.~\ref{sec:loss} elaborates training details.

\subsection{Preliminaries}
\label{sec:prem}
3D-GS \cite{kerbl3Dgaussians} represents the scene with a set of anisotropic 3D Gaussians that inherit the differential properties of volumetric representation while be efficiently rendered via a tile-based rasterization. 

Starting from a set of Structure-from-Motion (SfM) points, each point is designated as the position (mean) $\mu$ of a 3D Gaussian:
\begin{equation}
G(x) = e^{-\frac{1}{2} (x-\mu)^T \Sigma^{-1} (x-\mu)},
\label{gs}
\end{equation}
where $x$ is an arbitrary position within the 3D scene and $\Sigma$ denotes the covariance matrix of the 3D Gaussian. $\Sigma$ is formulated using a scaling matrix $S$ and rotation matrix $R$ to maintain its positive semi-definite:
\begin{equation}
\Sigma = RSS^TR^T,
\label{sigma}
\end{equation}
In addition to color $c$ modeled by Spherical harmonics, each 3D Gaussian is associated with an opacity $\alpha$ which is multiplied by $G(x)$ during the blending process. 

Distinct from conventional volumetric representations, 3D-GS efficiently renders the scene via tile-based rasterization instead of resource-intensive ray-marching. The 3D Gaussian $G(x)$ are first transformed to 2D Gaussians $G'(x)$ on the image plane following the projection process as described in \cite{zwicker2001ewa}. Then a tile-based rasterizer is designed to efficiently sort the 2D Gaussians and employ $\alpha$-blending: 
\begin{equation}
C(x') = \sum_{i \in N} c_i \sigma_i \prod_{j=1}^{i-1} (1 - \sigma_j), \quad \sigma_i = \alpha_i G'_{i}(x') ,
\label{gs_rendering}
\end{equation}
where $x'$ is the queried pixel position and N denotes the number of sorted 2D Gaussians associated with the queried pixel. Leveraging the differentiable rasterizer, all attributes of the 3D Gaussians are learnable and directly optimized end-to-end via training view reconstruction.

\subsection{\modelname}

\subsubsection{Anchor Point Initialization}
\label{sec:Anchor}
Consistent with existing methods~\cite{kerbl3Dgaussians,xu2022point}, we use the sparse point cloud from COLMAP~\cite{schoenberger2016sfm} as our initial input. 
We then voxelize the scene from this point cloud $\mathbf{P}\in\mathbb{R}^{M\times3}$ as:
\begin{equation}
    \label{eq:voxelization}
    \mathbf{V} = \left\{\left\lfloor\frac{\mathbf{P}}{\epsilon}\right\rceil\right\} \cdot \epsilon,
\end{equation}
where $\mathbf{V}\in\mathbb{R}^{N\times3}$ denotes voxel centers, and $\epsilon$ is the voxel size. We then remove duplicate entries, denoted by $\{\cdot\}$ to reduce the redundancy and irregularity in $\mathbf{P}$.

The center of each voxel $v\in\mathbf{V}$ is treated as an anchor point, equipped with a local context feature $f_v \in \mathbb{R}^{32}$, 
a scaling factor $l_v \in \mathbb{R}^{3}$, 
and $k$ learnable offsets $\mathbf{O}_v\in\mathbb{R}^{k\times3}$.
In a slight abuse of terminology, we will denote the anchor point as $v$ in the following context. We further enhance $f_v$ to be multi-resolution and view-dependent.
For each anchor $v$, we 
1) create a features bank $\{f_v, f_{v_{\downarrow_1}}, f_{v_{\downarrow_2}}\} $, where $\downarrow_n$ denotes $f_v$ being down-sampled by $2^n$ factors; 
2) blend the feature bank with view-dependent weights to form an integrated anchor feature $\hat{f_v}$. Specifically,
Given a camera at position $\mathbf{x}_c$ and an anchor at position $\mathbf{x}_v$, we calculate their relative distance and viewing direction with:
\begin{align}
\label{eq:rel_dist_view}
    \delta_{vc} = \lVert \mathbf{x}_v - \mathbf{x}_c \lVert_2,
    \vec{\mathbf{d}}_{vc} = \frac{\mathbf{x}_v - \mathbf{x}_c}{\lVert \mathbf{x}_v - \mathbf{x}_c \lVert_2},
\end{align} 
then weighted sum the feature bank with weights predicted from a tiny MLP $F_{w}$:
\begin{align}
    \{w, w_1, w_2\} &= \operatorname{Softmax}(F_{w}(\delta_{vc}, \vec{\mathbf{d}}_{vc})), \\
    \hat{f_v} &= w\cdot f_v + w_1\cdot f_{v_{\downarrow_1}} + w_2\cdot f_{v_{\downarrow_2}}, 
\end{align}

\subsubsection{Neural Gaussian Derivation}
\label{sec:Neural}
In this section, we elaborate on how our approach derives neural Gaussians from anchor points. 
Unless specified otherwise, $F_*$ represents a particular MLP throughout the section. Moreover, we introduce two efficient pre-filtering strategies to reduce MLP overhead.

We parameterize a neural Gaussian with its position $\mu\in\mathbb{R}^{3}$, opacity $\alpha\in\mathbb{R}$, covariance-related quaternion $q\in\mathbb{R}^4$ and scaling $s\in\mathbb{R}^3$, and color $c\in\mathbb{R}^3$.
As shown in Fig.~\ref{fig:pipeline}(b),  
for each visible anchor point within the viewing frustum, we spawn $k$ neural Gaussians and predict their attributes.
Specifically, given an anchor point located at $\mathbf{x}_v$, the positions of its neural Gaussians are calculated as:
\begin{equation}
    \{\mu_0,...,\mu_{k-1}\} = \mathbf{x}_v+\{\mathcal{O}_0,\dots,\mathcal{O}_{k-1}\}\cdot l_v,
\end{equation}
where $\{\mathcal{O}_0, \mathcal{O}_1, ...,\mathcal{O}_{k-1}\} \in \mathbb{R}^{k \times 3}$ are the learnable offsets and $l_v$ is the scaling factor associated with that anchor, as described in Sec.~\ref{sec:Anchor}.
The attributes of $k$ neural Gaussians are directly decoded from the anchor feature $\hat{f}_v$, the relative viewing distance $\delta_{vc}$ and direction $\vec{\mathbf{d}}_{vc}$ between the camera and the anchor point (Eq.~\ref{eq:rel_dist_view}) through individual MLPs, denoted as $F_{\alpha}$, $F_{c}$, $F_{q}$ and $F_{s}$.
Note that attributes are decoded in \emph{one-pass}. For example, opacity values of neural Gaussians spawned from an anchor point are given by:
\begin{equation}
    \{\alpha_0, ..., \alpha_{k-1}\} = F_{\alpha}(\hat{f_v}, \delta_{vc}, \vec{\mathbf{d}}_{vc}),
\end{equation}
their colors $\{c_i\}$, quaternions $\{q_i\}$ and scales $\{s_i\}$ are similarly derived.  
Implementation details are in supplementary.

Note that the prediction of neural Gaussian attributes are \emph{on-the-fly}, meaning that only anchors visible within the frustum are activated to spawn neural Gaussians. 
To make the rasterization more efficient, we only keep neural Gaussians whose opacity values are larger than a predefined threshold $\tau_\alpha$.
This substantially cuts down the computational load and helps our method maintain a high rendering speed on-par with the original 3D-GS. 

\subsection{Anchor Points Refinement}
\label{sec:intern}
\begin{figure}[t]
  \centering
   \includegraphics[width=\linewidth]{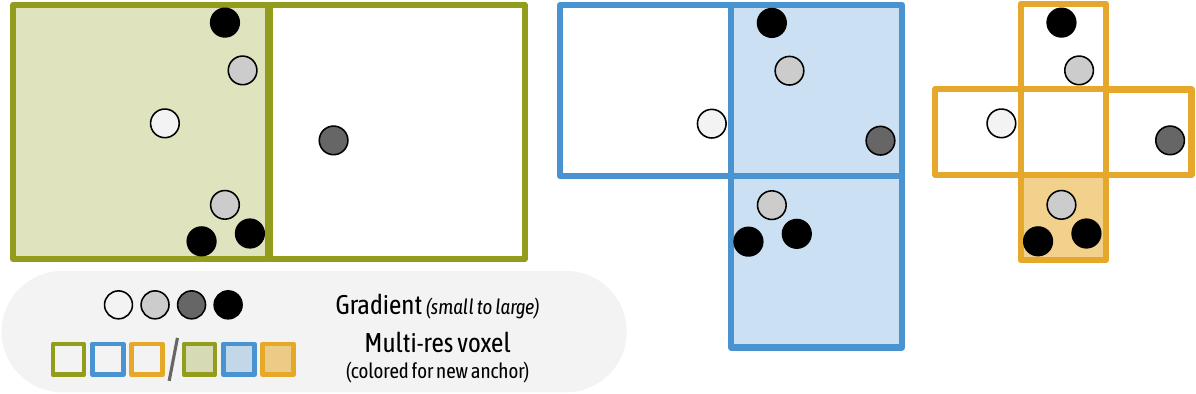}
   \caption{\textbf{Growing operation.} We develop an anchor growing policy guided by the gradients of the neural Gaussians. From left to right, we spatially quantize neural Gaussians into multi-resolution voxels ($m\in\{1,2,3\}$) of size $\{\epsilon_{g}^{(m)}\}$. New anchors are added to voxels with aggregated gradients larger than $\{\tau_{g}^{(m)}$\}.}
   \label{fig:growing}
\end{figure}

\paragraph{Growing Operation.}
Since neural Gaussians are closely tied to their anchor points which are initialized from SfM points, their modeling power is limited to a local region, as has been pointed out in~\citep{xu2022point,kerbl3Dgaussians}. This poses challenges to the initial placement of anchor points, especially in texture-less and less observed areas. 
We therefore propose an error-based anchor growing policy that grows new anchors where neural Gaussians find \emph{significant}. 
To determine a \emph{significant} area, we first spatially quantize the neural Gaussians by constructing voxels of size $\epsilon_{g}$. 
For each voxel, we compute the averaged gradients of the included neural Gaussians over $N$ training iterations, denoted as $\nabla_{g}$.
Then, voxels with $\nabla_{g} > \tau_{g}$ is deemed as \emph{significant}, where $\tau_{g}$ is a pre-defined threshold;
and a new anchor point is thereby deployed at the center of that voxel if there was no anchor point established.
Fig.~\ref{fig:growing} illustrates this growing operation.
In practice, we quantize the space into multi-resolution voxel grid to allow new anchors to be added at different granularity, where 
\begin{equation}
    \label{eq:hierachyparams}
        \epsilon_{g}^{(m)}=\epsilon_{g}/4^{m-1}, \quad \tau_{g}^{(m)}={\tau_{g}}*2^{m-1}, 
\end{equation}
where $m$ denotes the level of quantization.
To further regulate the addition of new anchors, we apply a random elimination to these candidates. This cautious approach to adding points effectively curbs the rapid expansion of anchors.

\paragraph{Pruning Operation}
To eliminate \emph{trivial} anchors, we accumulate the opacity values of their associated neural Gaussians over $N$ training iterations. If an anchor fails to produce neural Gaussians with a satisfactory level of opacity, we then remove it from the scene.

\subsection{Losses Design}
\label{sec:loss}
We optimize the learnable parameters and MLPs with respect to the $\mathcal{L}_1$ loss over rendered pixel colors, with SSIM term~\citep{1284395} $\mathcal{L}_{SSIM}$ and volume regularization~\citep{lombardi2021mixture} $\mathcal{L}_{vol}$. The total supervision is given by:
\begin{equation}
     \mathcal{L} =\mathcal{L}_1 + \lambda_{\text{SSIM}}\mathcal{L}_{\text{SSIM}} + \lambda_{\text{vol}}\mathcal{L}_{\text{vol}},
\end{equation}
where the volume regularization $\mathcal{L}_{\text{vol}}$ is:
\begin{equation}
     \mathcal{L}_{\text{vol}} = \sum_{i=1}^{N_{\text{ng}}} \operatorname{Prod}(s_i).
\end{equation}
Here, $N_{\text{ng}}$ denotes the number of neural Gaussians in the scene and $\text{Prod}(\cdot)$ is the product of the values of a vector, \eg, in our case the scale $s_i$ of each neural Gaussian. The volume regularization term encourages the neural Gaussians to be small with minimal overlapping.
\section{Experiments}
\label{sec:exp}

\begin{table*}[]
\centering
\caption{\textbf{Quantitative comparison to previous methods on real-world datasets.} 
Competing metrics are extracted from respective papers.
}
\label{tab:quality}
\resizebox{0.86\linewidth}{!}{
\begin{tabular}{c|ccc|ccc|ccc}
\toprule
Dataset & \multicolumn{3}{c|}{Mip-NeRF360} & \multicolumn{3}{c|}{Tanks\&Temples} & \multicolumn{3}{c}{Deep Blending} \\
\begin{tabular}{c|c} Method & Metrics \end{tabular}  & PSNR \(\uparrow\) & SSIM \(\uparrow\) & LPIPS \(\downarrow\) & PSNR \(\uparrow\) & SSIM \(\uparrow\) & LPIPS \(\downarrow\) & PSNR \(\uparrow\) & SSIM \(\uparrow\) & LPIPS \(\downarrow\) \\
\midrule
\textbf{3D-GS}~\cite{kerbl3Dgaussians} & \cellcolor{tabthird}28.69 & \cellcolor{tabfirst}0.870 &\cellcolor{tabfirst} 0.182 & \cellcolor{tabsecond}23.14 & \cellcolor{tabsecond}0.841 & \cellcolor{tabsecond}0.183 & \cellcolor{tabsecond}29.41 & \cellcolor{tabsecond}0.903 & \cellcolor{tabfirst}0.243 \\
\textbf{Mip-NeRF360} ~\cite{barron2022mipnerf360} & \cellcolor{tabfirst}29.23 & \cellcolor{tabthird}0.844 & \cellcolor{tabsecond}0.207 & \cellcolor{tabthird}22.22 & \cellcolor{tabthird}0.759 & \cellcolor{tabthird}0.257 & \cellcolor{tabthird}29.40 & \cellcolor{tabthird}0.901 & \cellcolor{tabsecond}0.245 \\
\textbf{iNPG} ~\cite{muller2022instant} & 26.43 & 0.725 & 0.339 & 21.72 & 0.723 & 0.330 & 23.62 & 0.797 & 0.423 \\
\textbf{Plenoxels} ~\cite{yu2022plenoxels} & 23.62 & 0.670 & 0.443 & 21.08 & 0.719 & 0.379 & 23.06 & 0.795 & 0.510 \\ \hline
\textbf{Ours} & \cellcolor{tabsecond}28.84 & \cellcolor{tabsecond}0.848 & \cellcolor{tabthird}0.220 & \cellcolor{tabfirst}23.96 &\cellcolor{tabfirst} 0.853 &\cellcolor{tabfirst} 0.177 & \cellcolor{tabfirst}30.21 & \cellcolor{tabfirst}0.906 & \cellcolor{tabthird}0.254 \\
\bottomrule
\end{tabular}}
\end{table*}

\begin{table}[]
\centering
\caption{\textbf{Performance comparison.} Rendering FPS and storage size are reported. Storage size reduction ratio is indicated by ($\downarrow$). Rendering speed of both methods are measured on our machine.
}
\label{tab:performance}
\resizebox{\linewidth}{!}{
\begin{tabular}{c|cc|cc|cc}
\toprule
Dataset & \multicolumn{2}{c|}{Mip-NeRF360} & \multicolumn{2}{c|}{Tanks\&Temples} & \multicolumn{2}{c}{Deep Blending}  \\
 & FPS & Mem (MB) & FPS & Mem (MB) & FPS & Mem (MB) \\ 
\midrule
\textbf{3D-GS} & 97 & 693 & \textbf{123} & 411 & 109 & 676 \\
\textbf{Ours} & \textbf{102} & \textbf{156} (4.4$\times$ $\downarrow$) & 110 & \textbf{87} (4.7$\times$ $\downarrow$) & \textbf{139} & \textbf{66} (10.2$\times$ $\downarrow$) \\
\bottomrule

\end{tabular}

}
\end{table}

\subsection{Experimental Setup}
\paragraph{Dataset and Metrics.} 
We conducted a comprehensive evaluation across 27 scenes from publicly available datasets. 
Specifically, we tested our approach on all available scenes tested in the 3D-GS~\cite{kerbl3Dgaussians}, including seven scenes from Mip-NeRF360~\cite{barron2022mipnerf360}, two scenes from Tanks\&Temples~\cite{Knapitsch2017}, two scenes from DeepBlending~\cite{hedman2018deep} and synthetic Blender dataset~\cite{mildenhall2021nerf}. We additionally evaluated on datasets with contents captured at multiple LODs to demonstrate our advantages in view-adaptive rendering. 
Six scenes from BungeeNeRF~\cite{xiangli2022bungeenerf} and two scenes from VR-NeRF~\cite{VRNeRF} are selected. The former provides multi-scale outdoor observations and the latter captures intricate indoor environments. Apart from the commonly used metrics (PSNR, SSIM~\cite{1284395}, and LPIPS~\cite{Zhang_2018_CVPR}), we additionally report the storage size (MB) and the rendering speed (FPS) for model compactness and performance efficiency. We provide the averaged metrics over all scenes of each dataset in the main paper and leave the full quantitative results on each scene in the supplementary.

\paragraph{Baseline and Implementation.}
3D-GS~\cite{kerbl3Dgaussians} is selected as our main baseline for its established SOTA performance in novel view synthesis.
Both 3D-GS and our method were trained for $30$k iterations. 
We also record the results of Mip-NeRF360~\cite{barron2022mipnerf360}, iNGP~\cite{muller2022instant} and Plenoxels~\cite{yu2022plenoxels} as in \cite{kerbl3Dgaussians}.

For our method, we set $k=10$ for all experiments. All the MLPs employed in our approach are $2$-layer MLPs with ReLU activation; the dimensions of the hidden units are all $32$. For anchor points refinement, we average gradients over $N=100$ iterations, and by default use $\tau_{g}=64\epsilon$. On intricate scenes and the ones with dominant texture-less regions, we use $\tau_{g}=16\epsilon$. An anchor is pruned if the accumulated opacity of its neural Gaussians is less than $0.5$ at each round of refinement. The two loss weights $\lambda_{\text{SSIM}}$ and $\lambda_{\text{vol}}$ are set to $0.2$ and $0.001$ in our experiments. 
Please check the supplementary material for more details.

\subsection{Results Analysis}

\begin{figure*}[t!]
	\centering
	\includegraphics[width=\linewidth]{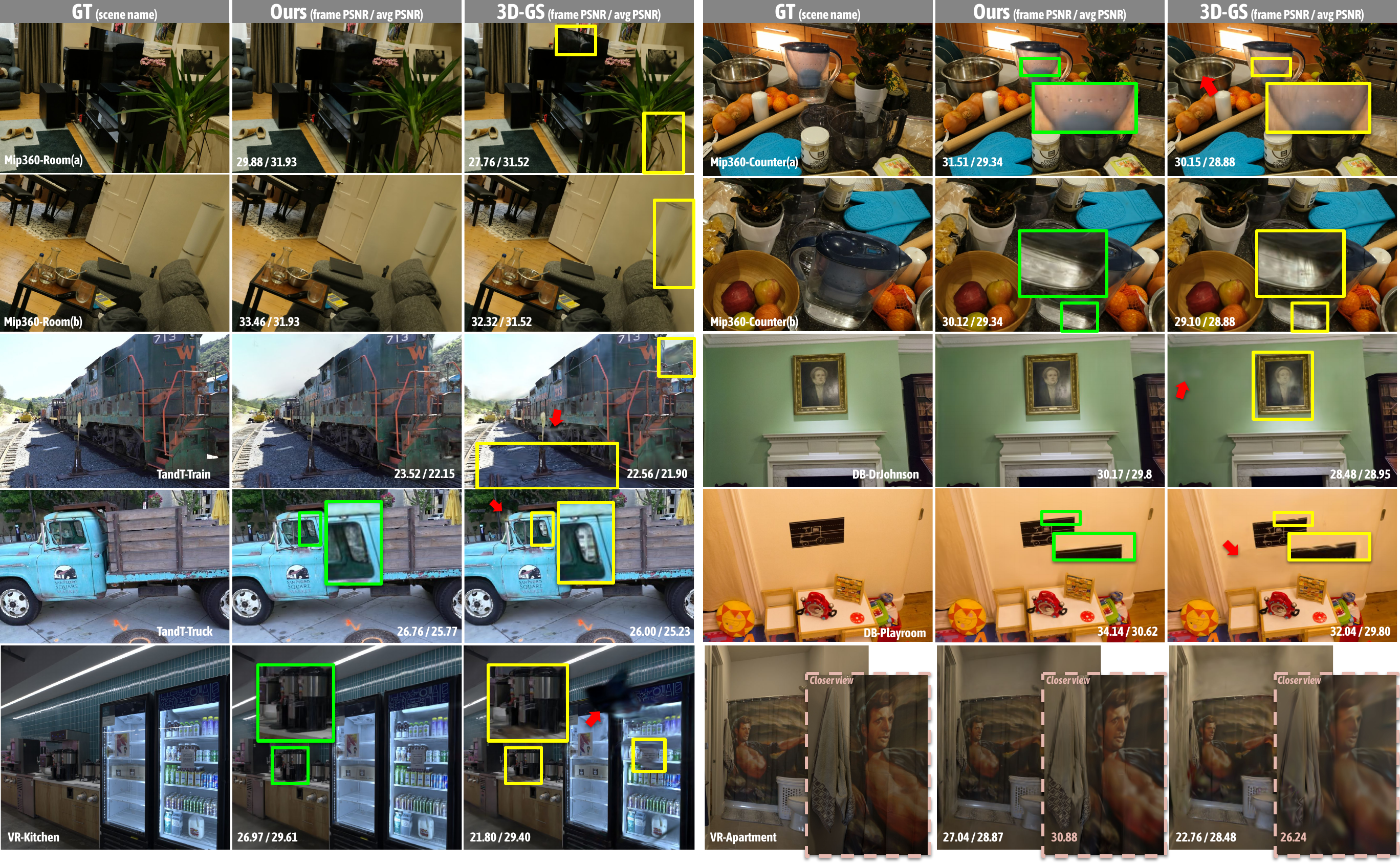}
	\vspace{-20pt}
	\caption{\textbf{Qualitative comparison of \modelname and 3D-GS~\citep{kerbl3Dgaussians} across diverse datasets~\citep{barron2022mipnerf360,Knapitsch2017,DeepBlending2018,VRNeRF}.} Patches that highlight the visual differences are emphasized with \textcolor{red}{arrows} and \textcolor{green}{green} \& \textcolor{yellow}{yellow} insets for clearer visibility. Our approach constantly outperforms 3D-GS on these scenes, with evident advantages in challenging scenarios,~\eg thin geometry and fine-scale details (\textsc{Mip360-Room}(a), \textsc{Mip360-Counter}(a)), texture-less regions (\textsc{DB-DrJohnson}, \textsc{DB-Playroom}), light effects (\textsc{Mip360-Counter}(b), \textsc{DB-DrJohnson}), insufficient observations (\textsc{TandT-Train}, \textsc{VR-Kitchen}). It can also be observed~(\eg \textsc{VR-Apartment}) that our model is superior in representing contents at varying scales and viewing distances.}
	\vspace{-10pt}
	\label{fig:results}
\end{figure*}

Our evaluation was conducted on diverse datasets, ranging from synthetic object-level scenes, indoor and outdoor environments, to large-scale urban scenes and landscapes. 
A variety of improvements can be observed especially on challenging cases, such as texture-less area, insufficient observations, fine-scale details and view-dependent light effects. See Fig.~\ref{fig:teaser} and Fig.~\ref{fig:results} for examples. 

\paragraph{Comparisons.} In assessing the quality of our approach, we compared with 3D-GS~\citep{kerbl3Dgaussians}, Mip-NeRF360~\citep{barron2022mipnerf360}, iNGP~\citep{muller2022instant} and Plenoxels~\citep{yu2022plenoxels} on real-world datasets. Qualitative results are presented in Tab.~\ref{tab:quality}. 
The quality metrics for Mip-NeRF360, iNGP and Plenoxels align with those reported in the 3D-GS study. It can be noticed that our approach achieves comparable results with the SOTA algorithms on Mip-NeRF360 dataset, and surpassed the SOTA on Tanks\&Temples and DeepBlending, which captures more challenging environments with the presence of~\eg changing lighting, texture-less regions and reflections. In terms of efficiency, we evaluated rendering speed and storage size of our method and 3D-GS, as shown in Tab.~\ref{tab:performance}. Our method achieved real-time rendering while using less storage, indicating that our model is more compact than 3D-GS without sacrificing rendering quality and speed.
Additionally, akin to prior grid-based methods, our approach converged faster than 3D-GS. See supplementary material for more analysis. 

We also examined our method on the synthetic Blender dataset, which provides an exhaustive set of views capturing objects at $360^{\circ}$. A good set of initial SfM points is not readily available in this dataset, thus we start from 
$100k$ grid points and learn to grow and prune points with our anchor refinement operations. 
After $30$k iterations, we used the remained points as initialized anchors and re-run our framework.
The PSNR score and storage size compared with 3D-GS are presented in Tab.~\ref{tab:lod}. Fig.~\ref{fig:teaser} also demonstrates that our method can achieve better visual quality with more reliable geometry and texture details.

\begin{figure}[t]
  \centering
   \includegraphics[width=\linewidth]{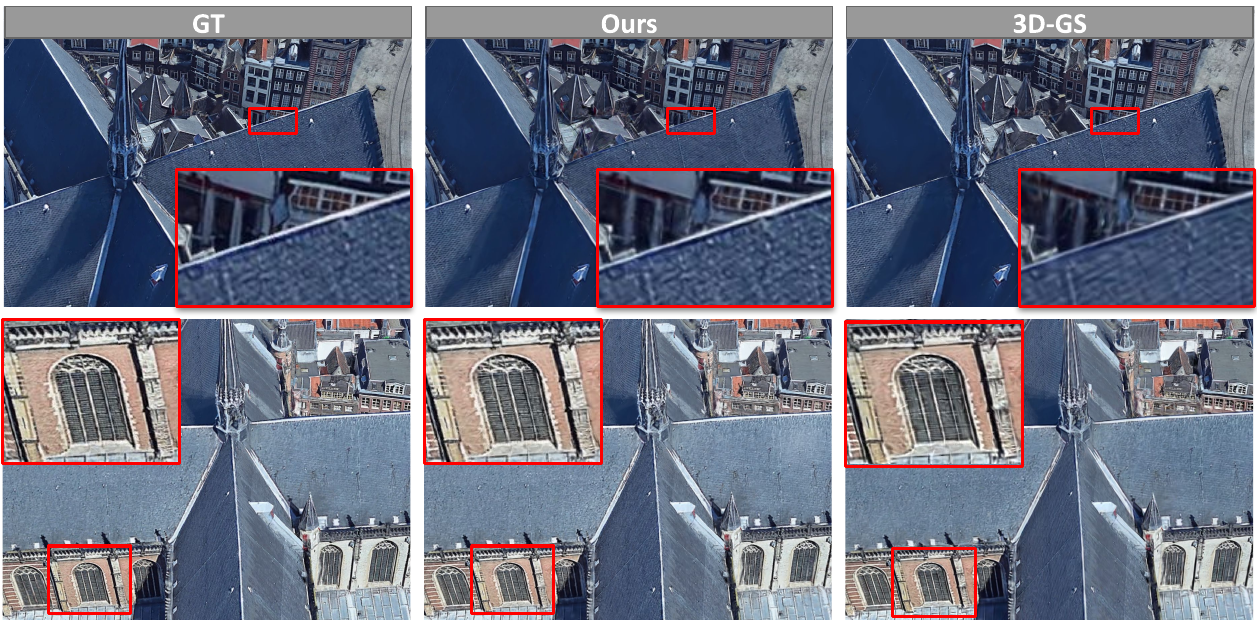}
   \caption{\textbf{Comparison on multi-scale scenes (w/ zoom-in cases).} We showcase the rendering outcomes at an unsceen closer scale on the \textsc{Amsterdam} scene from BungeeNeRF. Our method smoothly extrapolates to new viewing distances using refined neural Gaussian properties, remedying the needle-like artifacts of original 3D-GS caused by fixed Gaussian scaling values.}

   \label{fig:google}
\end{figure}

\paragraph{Multi-scale Scene Contents.} 
We examined our model's capability in handling multi-scale scene details on the BungeeNeRF and VR-NeRF datasets. As shown in Tab.~\ref{tab:lod}, our method achieved superior quality whilst using fewer storage size to store the model compared to 3D-GS \cite{kerbl3Dgaussians}. As illustrated in Fig.~\ref{fig:results} and Fig.~\ref{fig:google}, our method was superior in accommodating varying levels of detail in the scene. In contrast, images rendered from 3D-GS often suffered from noticeable blurry and needle-shaped artifacts. 
This is likely because that Gaussian attributes are optimized to overfit multi-scale training views, 
creating excessive Gaussians that work for each observing distance.
However, it can easily lead to ambiguity and uncertainty when synthesizing novel views,
since it lacks the ability to reason about viewing angle and distance. 
On contrary, our method efficiently encoded local structures into compact neural features, enhancing both rendering quality and convergence speed.
Details are provided in the supplementary material.

\begin{table}[]
\centering
\caption{\textbf{Qualitative comparison.} Our method is able to handle large-scale scenes~(\eg \textsc{BungeeNeRF}) with light-weight representation. Our method shows consistent compactness and effectiveness in complex lighting conditions and synthetic scenes.}
\vspace{-2pt}
\label{tab:lod}
\resizebox{\linewidth}{!}{
\begin{tabular}{c|cc|cc|cc}
\toprule
Dataset & \multicolumn{2}{c|}{BungeeNeRF} & \multicolumn{2}{c|}{VR-NeRF} & \multicolumn{2}{c}{Synthetic Blender} \\
 & PSNR & Mem (MB) & PSNR & Mem (MB) & PSNR & Mem (MB) \\
\midrule
\textbf{3D-GS} & 24.89 & 1606 & 28.94 & 263 & 33.32 & 53 \\
\textbf{Ours} & \textbf{27.01} & \textbf{203} (7.9$\times$ $\downarrow$) & \textbf{29.24} & \textbf{69} (3.8$\times$ $\downarrow$) & \textbf{33.68} & \textbf{14} (3.8$\times$ $\downarrow$) \\
\bottomrule
\end{tabular}
}
\end{table}

\begin{figure}[t]
  \centering
   \includegraphics[width=\linewidth]{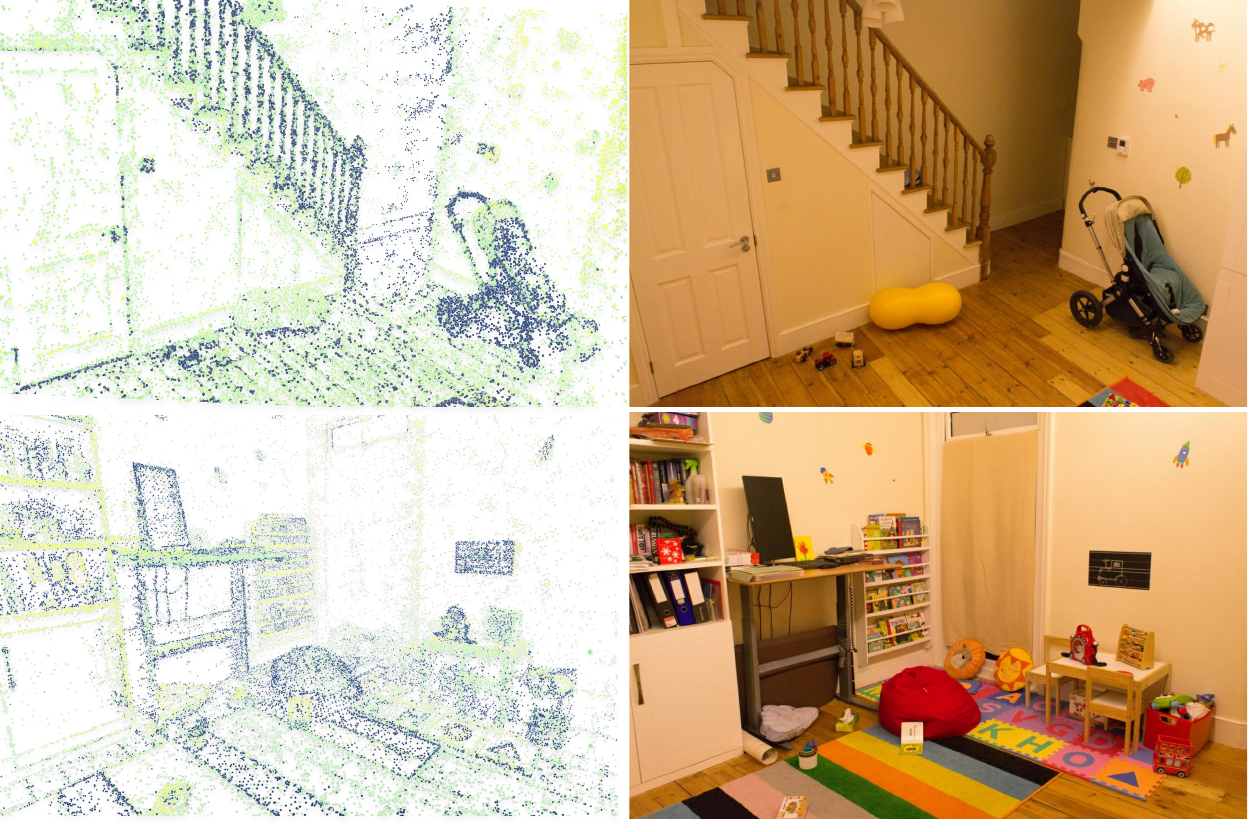}
   
   \caption{\textbf{Anchor feature clustering.} 
   We cluster anchor features (\textsc{DB-Playroom}) into $3$ clusters using K-means~\citep{krishna1999genetic} and visualize the result. The clustered features show clues of scene contents,~\eg the banister, stroller, desk and monitor can be clearly identified. Anchors on the wall and floor are also respectively grouped together. This shows that our approach improves the interpretability of 3D-GS model, and has the potential to be scaled-up on much larger scenes exploiting reusable features.}
   \label{fig:clustering}
\end{figure}

\begin{figure}[t]
  \centering
   \includegraphics[width=\linewidth]{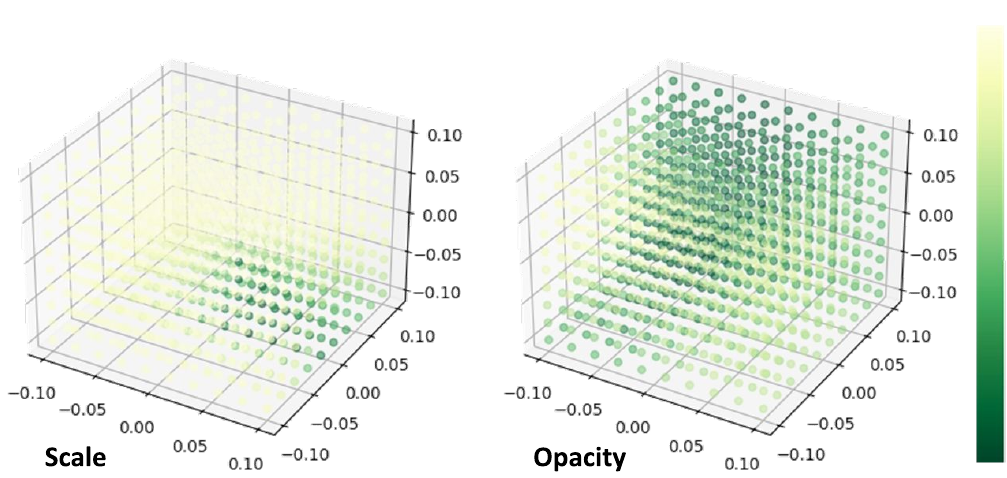}
   \caption{\textbf{View-adaptive neural Gaussian attributes.} We visualize the decoded attributes of a \emph{single} neural Gaussian observed at different positions. Each point corresponds to a viewpoint in space. The color of the point denotes the intensity of attributes decoded for this view (left: $F_s \rightarrow s_i$; right: $F_\alpha \rightarrow \alpha_i$). 
   This pattern indicates that attributes of a neural Gaussian adapt to viewpoint changing, while exhibiting a certain degree of local continuity.}
   \label{fig:view}
\end{figure}

\paragraph{Feature Analysis.} 
We further perform an analysis of the learnable anchor features and the selector mechanism.
As depicted in Fig.~\ref{fig:clustering}, 
the clustered pattern suggests that the compact anchor feature spaces adeptly capture regions with similar visual attributes and geometries, as evidenced by their proximity in the encoded feature space.
\paragraph{View Adaptability.}
To support that our neural Gaussians are view-adaptive, we explore how the values of attributes change when the same Gaussian is observed from different positions. Fig.~\ref{fig:view} demonstrates a varying distribution of attributes intensity at different viewing positions, while maintaining a degree of local continuity. This accounts for the superior \emph{view adaptability} of our method compared to the static attributes of 3D-GS, as well as its enhanced generalizability to novel views.

\begin{figure}[t]
  \centering
   \includegraphics[width=\linewidth]{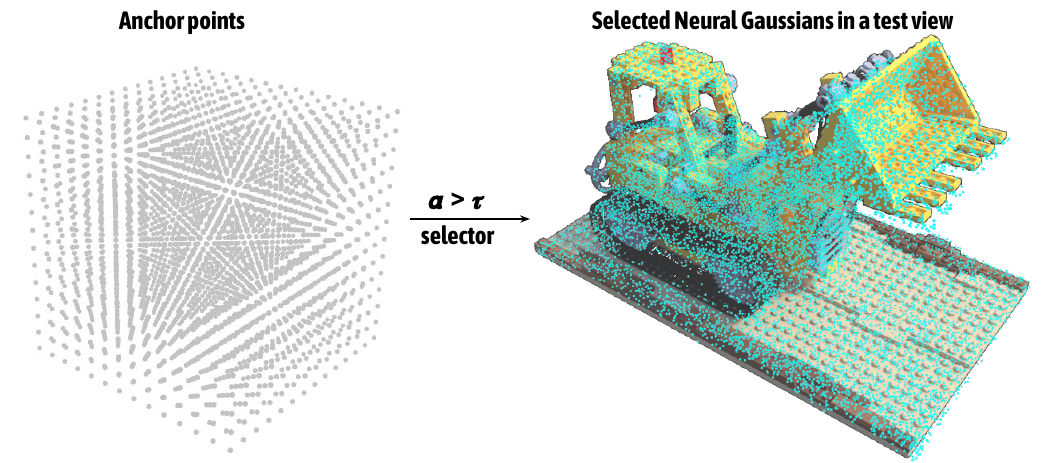}
    \caption{\textbf{Geometry culling via selector.} (Left) Anchor points from randomly initialized points; (Right) Activated neural Gaussians derived from each anchor under the current view. In synthetic Blender scenes, with all 3D Gaussians visible in the viewing frustum, our opacity filtering functions similar to a geometry proxy estimator, excluding unoccupied regions before rasterization.}
   \label{fig:selector_lego}
\end{figure}

\paragraph{Selection Process by Opacity.}
We examine the decoded opacity from neural Gaussians and our opacity-based selection process (Fig.~\ref{fig:pipeline}(b)) from two aspects.
First, without the anchor point refinement module, we filter neural Gaussian using their decoded opacity values to extract geometry from a random point cloud. Fig.~\ref{fig:selector_lego} demonstrates that the remained neural Gaussians effectively reconstruct the coarse structure of the bulldozer model from random points, highlighting its capability for implicit geometry modeling under mainly rendering-based supervision.
We found this is conceptually similar to the proposal network utilized in \cite{barron2022mipnerf360}, serving as the geometry proxy estimator for efficient sampling.

\begin{figure}[t]
  \centering
   \includegraphics[width=\linewidth]{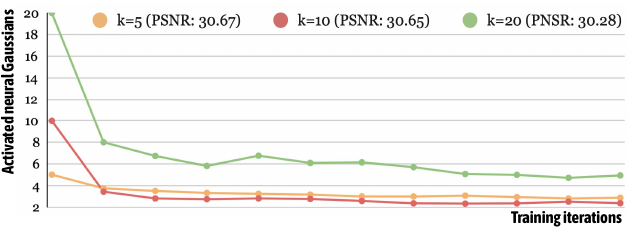}
   \vspace{-15pt}
   \caption{\textbf{Learning with different $k$ values.} 
   Despite varying initial $k$ values under different hyper-parameter settings, they converge to activate a similar number of neural Gaussians with comparable rendering fidelity.
   }
   \label{fig:different_k}
\end{figure}

Second, we apply different $k$ values in our method. Fig.~\ref{fig:different_k} shows that regardless of the $k$ value, the final number of activated neural Gaussians converges to a similar amount through the training, indicating \modelname's preference to select a collection of non-redundant Gaussians that are sufficient to represent the scene.

\subsection{Ablation Studies}

\begin{table}[]
\centering
\caption{\textbf{Effects of filtering}. \textsc{Filter 1} refers to selecting anchors by view frustum and \textsc{Filter 2} refers to the opacity-based selection process. The filtering method has no notable impact on fidelity, but greatly affects inference speed.}
\vspace{-2pt}
\label{tab:Filter Strategies}
\resizebox{0.9\linewidth}{!}{
\begin{tabular}{c|cc|cc}
\toprule
  Scene & \multicolumn{2}{c|}{\textsc{DB-Playroom}} & \multicolumn{2}{c}{\textsc{DB-DrJohnson}}  \\
\begin{tabular}{cc} & \end{tabular} & PSNR & FPS & PSNR & FPS \\ \midrule
\textsc{No Filters} & 30.4           & 84          & 29.7           & 79 \\
\textsc{Filter 1} & 30.3           & 118          & 29.6           & 100\\
\textsc{Filter 2}          & 30.6           & 109          & 29.7           & 104 \\ \hline
\textsc{Full}
& 30.62          & 150         & 29.8           & 129 \\
\bottomrule
\end{tabular}
}
\end{table}

\paragraph{Efficacy of Filtering Strategies.} 
We evaluated our filtering strategies (Sec.~\ref{sec:Neural}), which we found crucial for speeding up our method. As Tab.~\ref{tab:Filter Strategies} shows, while these strategies had no notable effect on fidelity, they significantly enhanced inference speed. However, there was a risk of masking pertinent neural Gaussians, which we aim to address in future works.

\paragraph{Efficacy of Anchor Points Refinement Policy.} We evaluated our growing and pruning operations described in Sec.~\ref{sec:intern}. Tab.~\ref{tab:intern} shows the results of disabling each operation in isolation and maintaining the rest of the method. We found that the addition operation is crucial for accurately reconstructing details and texture-less areas,
while the pruning operation plays an important role in eliminating trivial Gaussians and maintaining the efficiency of our approach.

\begin{table}[]
\centering
\caption{\textbf{Anchor refinement}. The growing operation is essential for fidelity since it improves the poor initialization. The pruning operation controls the increasing of storage size and optimizes the quality of remained anchors.}
\vspace{-2pt}
\label{tab:intern}
\resizebox{\linewidth}{!}{
\begin{tabular}{c|cc|cc}
\toprule
  Scene & \multicolumn{2}{c|}{\textsc{DB-Playroom}} & \multicolumn{2}{c}{\textsc{DB-DrJohnson}}  \\
\begin{tabular}{cc} & \end{tabular} & PSNR & Mem (MB) & PSNR & Mem (MB) \\ \midrule
\textsc{None}              & 28.45         & 24         & 28.81          & 12         \\
\textsc{w/ Pruning}      & 29.12         & 23         & 28.51          & 12         \\
\textsc{w/ Growing}      & 30.54         & 71         & 29.75          & 76         \\ \hline
\textsc{Full}              & 30.62         & 63         & 29.80          & 68         \\
\bottomrule
\end{tabular}
}
\end{table}

\subsection{Discussions and Limitations}

Through our experiments, we found that the initial points play a crucial role for high-fidelity results.
Initializing our framework from SfM point clouds is a swift and viable solution, considering these point clouds usually arise as a byproduct of image calibration processes. However, this approach may be suboptimal for scenarios dominated by large texture-less regions.
Despite our anchor point refinement strategy can remedy this issue to some extent, it still suffers from extremely sparse points.
We expect that our algorithm will progressively improve as the field advances, yielding more accurate results. 
Further details are discussed in the supplementary material.
\section{Conclusion}
\label{sec:conclusion}

In this work, we introduce \modelname, a novel 3D neural scene representation for efficient view-adaptive rendering.
The core of \modelname lies in its structural arrangement of 3D Gaussians guided by anchor points from SfM, whose attributes are on-the-fly decoded from view-dependent MLPs. 
We show that our approach leverages a much more compact set of Gaussians to achieve comparable or even better results than the SOTA algorithms. 
The advantage of our \emph{view-adaptive} neural Gaussians is particularly evident in challenging cases where 3D-GS usually fails. 
We further show that our anchor points encode local features in a meaningful way that exhibits semantic patterns to some degree, suggesting its potential applicability in a range of versatile tasks such as large-scale modeling, manipulation and interpretation in the future.  
% The core of \modelname lies in its dual-layered hierarchical structure, augmented by view-dependent MLP-based predictors. 
% This innovative design enables the efficient rendering of neural Gaussians, facilitating adaptation to inconsistent viewing angles. 
% Our comprehensive comparisons have established that \modelname not only excels in adapting to diverse scenes but also achieves rendering quality at par with the current state-of-the-art. In terms of performance, our method realizes real-time rendering and reduces storage size compared to other Gaussian-based methods thanks to our lightweight architecture. 

{
    \small
    \bibliographystyle{ieeenat_fullname}
    \bibliography{main}
}

 \clearpage
\setcounter{page}{1}
\maketitlesupplementary

% \section{Suppleme}
% \label{sec:rationale}
% % 
% Having the supplementary compiled together with the main paper means that:
% % 
% \begin{itemize}
% \item The supplementary can back-reference sections of the main paper, for example, we can refer to \cref{sec:intro};
% \item The main paper can forward reference sub-sections within the supplementary explicitly (e.g. referring to a particular experiment); 
% \item When submitted to arXiv, the supplementary will already included at the end of the paper.
% \end{itemize}
% % 
% To split the supplementary pages from the main paper, you can use \href{https://support.apple.com/en-ca/guide/preview/prvw11793/mac#:~:text=Delete%20a%20page%20from%20a,or%20choose%20Edit%20%3E%20Delete).}{Preview (on macOS)}, \href{https://www.adobe.com/acrobat/how-to/delete-pages-from-pdf.html#:~:text=Choose%20%E2%80%9CTools%E2%80%9D%20%3E%20%E2%80%9COrganize,or%20pages%20from%20the%20file.}{Adobe Acrobat} (on all OSs), as well as \href{https://superuser.com/questions/517986/is-it-possible-to-delete-some-pages-of-a-pdf-document}{command line tools}.
\section{Overview}
This supplementary is organized as follows: 
(1) In the first section, we elaborate implementation details of our \modelname, including anchor point feature enhancement (Sec.3.2.1), structure of MLPs (Sec.3.2.2) and anchor point refinement strategies (Sec.3.3);
% (Sec.~\ref{sec:Anchor}), structure of MLPs (Sec.~\ref{sec:Neural}) and anchor point refinement strategies (Sec.~\ref{sec:intern});
(2) The second part describes our dataset preparation steps. We then show additional experimental results and analysis based on our training observations. % and provide more comparisons against SOTA method. 
% (3) Finally, we discuss potential improvements that can be made in the future and present limitations of \emph{\modelname}.

% We also include supplementary videos that highlight our results and draw comparisons with the original 3D-GS. These videos are categorized and stored as follows:
% \begin{itemize}
% \item The \textsc{Scaffold-GS} folder contains five videos demonstrating our method's application on various scenes.
% \item The \textsc{comparison} folder contains three videos that compare our results with those of 3D-GS, illustrating the differences and advantages of our approach. \modelname is more consistent across views, has fewer floaters, and depicts details more clearly.
% \end{itemize}
% \LX{add some keyword/observation for the key difference?}

% \LM{LOD + surrounding}
% \begin{itemize}
%     \item 5-10 scenes with smooth trajectory
%     \item 2-3 comparison with baselines (side-by-side)
%     \item 1 scene with flip-test (interwined frames)
% \end{itemize}

\section{Implementation details.}
\label{sec:simplementation}
\begin{figure}[h]
  \centering
   \includegraphics[width=0.8\linewidth]{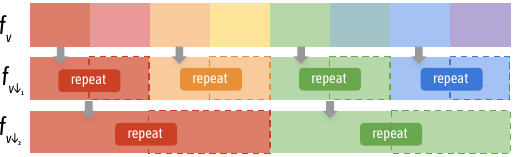}
   \caption{\textbf{Generation of Feature Bank.} We expand the anchor feature $f$ into a set of \emph{multi-resolution} features $\{f_{v}, f_{v_{\downarrow_1}}, f_{v_{\downarrow_2}}\}$ via slicing and repeating. This operation improves \modelname's ability to capture different scene granularity.}
   % From an anchor feature $f$, three features are generated by slicing and repeating.}
   \label{fig:feature}
\end{figure}

\begin{figure}[h]
  \centering
   \includegraphics[width=\linewidth]{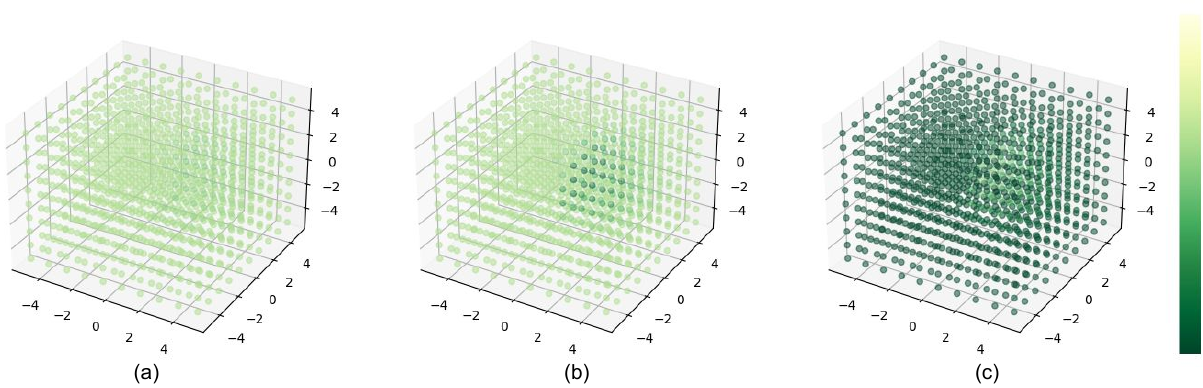}
   \caption{\textbf{View-based feature bank's weight distribution.} (a), (b) and (c) denote the predicted weights $\{w_2,w_1,w\}$ for $\{f_{v_{\downarrow_2}},f_{v_{\downarrow_1}},f_{v}\}$ from a group of uniformally distributed viewpoints. Light color denotes larger weights. For this anchor, finer features are more activated at center view positions.
   The patterns exhibit the ability to capture different scene granularities based on view direction and distance.}
   \label{fig:viewenc}
\end{figure}

\paragraph{Feature Bank.} To enhance the view-adaptability, we update the anchor feature through a view-dependent encoding. Following calculating the relative distance $\delta_{vc}$ and viewing direction $\vec{\mathbf{d}}_{vc}$ of a camera and an anchor, we predict a weight vector $w\in\mathbb{R}^3$ as follows:
\begin{equation}
    (w, w_1, w_2) = \operatorname{Softmax}(F_{w}(\delta_{vc}, \vec{\mathbf{d}}_{vc})),
\end{equation}
\noindent
where $F_{w}$ is a tiny MLP that serves as a view encoding function. We then encode the view direction information to the anchor feature $f_v$ by compositing a feature bank containing information with different resolutions as follows: 
\begin{equation}
    \hat{f_v} = w\cdot f_v + w_1\cdot f_{v_{\downarrow_1}} + w_2\cdot f_{v_{\downarrow_2}},
\end{equation}
In practice, we implement the feature bank via slicing and repeating, as illustrated in Fig.~\ref{fig:feature}.
We found this slicing and mixture operation improves \modelname's ability to capture different scene granularity. The distribution of feature bank's weights is illustrated in Fig.~\ref{fig:viewenc}.

\paragraph{MLPs as feature decoders.}

\begin{figure}[h]
  \centering
   \includegraphics[width=1.0\linewidth]{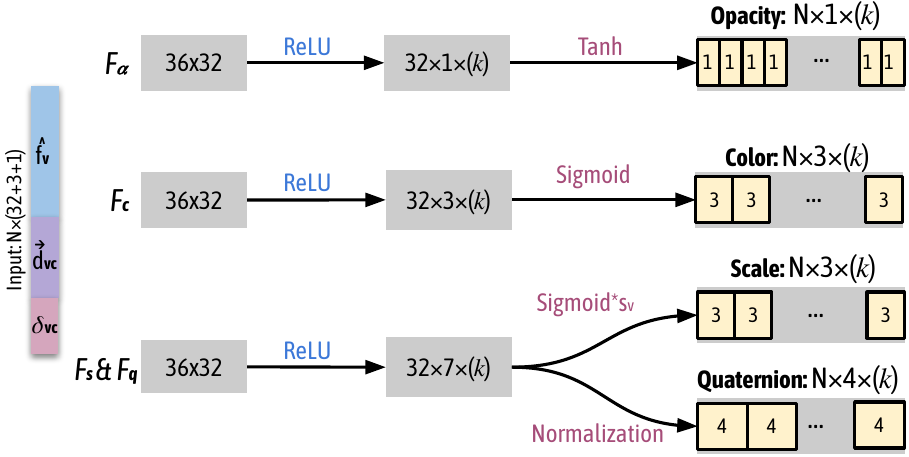}
   \caption{\textbf{MLP Structures.} 
   For each anchor point, we use small MLPs ($F_\alpha, F_c, F_s, F_q$) to predict attributes (opacity, color, scale and quaternion) of $k$ neural Gaussians. The input to MLPs are anchor feature $\hat{f_v}$, relative viewing direction $\vec{\mathbf{d}}_{vc}$ and distance $\delta_{vc}$ between the camera and anchor point. } 
   % link in \href{https://docs.google.com/drawings/d/1utWGdQCQohNucAcJe5gTIevLVDnuTnokImYviEBfmic/edit?usp=sharing}{googledraws}, beautify it as you like.}
   % From an anchor feature $f$, three features are generated by slicing and repeating.}
   \label{fig:mlps}
\end{figure}

The core MLPs include the opacity MLP $F_{\alpha}$, the color MLP $F_{c}$ and the covariance MLP $F_{s}$ and $F_{q}$. All of these $F_*$ are implemented in a \textsc{Linear $\rightarrow$ ReLU $\rightarrow$ Linear} style with the hidden dimension of $32$, as illustrated in Fig.~\ref{fig:mlps}. Each branch's output is activated with a head layer. 

\begin{itemize}
    \item For \emph{opacity}, the output is activated by $\operatorname{Tanh}$, where value 0 serves as a natural threshold for selecting valid samples and the final valid values can cover the full range of [0,1). 
    \item For \emph{color}, we activate the output with $\operatorname{Sigmoid}$ function:
        \begin{equation}
            \{c_0, ..., c_{k-1}\}=\operatorname{Sigmoid}(F_{c}),
        \end{equation}
        which constrains the color into a range of (0,1).
    \item For \emph{rotation}, we follow 3D-GS~\cite{kerbl3Dgaussians} and activate it with a normalization to obtain a valid quaternion. 
    \item For \emph{scaling}, we adjust the base scaling $s_v$ of each anchor with the MLP output as follows:
    \begin{equation}
        \{s_0, ..., s_{k-1}\}=\operatorname{Sigmoid}(F_{s})\cdot s_v,
    \end{equation}
\end{itemize}

% \begin{itemize}
%     \item Opacity: $\{\alpha_0, ..., \alpha_{k-1}\} = Tanh(F_{\alpha}),$
% \end{itemize}

% $F_{\alpha}$

\paragraph{Voxel Size.}

% The voxel size $\epsilon$ determines the finest resolution of the anchors. Considering the diverse scene complexity, we design two voxel size strategies: 1) Sort all initial points' 1 nearest-neighbor distance and choose the median value as the voxel size; 2) Manually set the $\epsilon=0.005$ or $\epsilon=0.01$. The former is adaptive to the density of initial point cloud and tends to produce denser anchors, which enjoys a better rendering quality at the risk of higher overhead. The manual manner performs well in most cases but may lose details in texture-less scenes. During experiments, the two strategies cover all the scenes well.

The voxel size $\epsilon$ sets the finest anchor resolution. We employ two strategies: 1) Use the median of the nearest-neighbor distances among all initial points: $\epsilon$ is adapted to point cloud density, yielding denser anchors with enhanced rendering quality but might introduce more computational overhead; 2) Set $\epsilon$ manually to either 0.005 or 0.01: this is effective in most scenarios but might lead to missing details in texture-less regions. We found these two strategies adequately accommodate various scene complexities in our experiments.

\begin{figure}[t]
  \centering
   \includegraphics[width=1.0\linewidth]{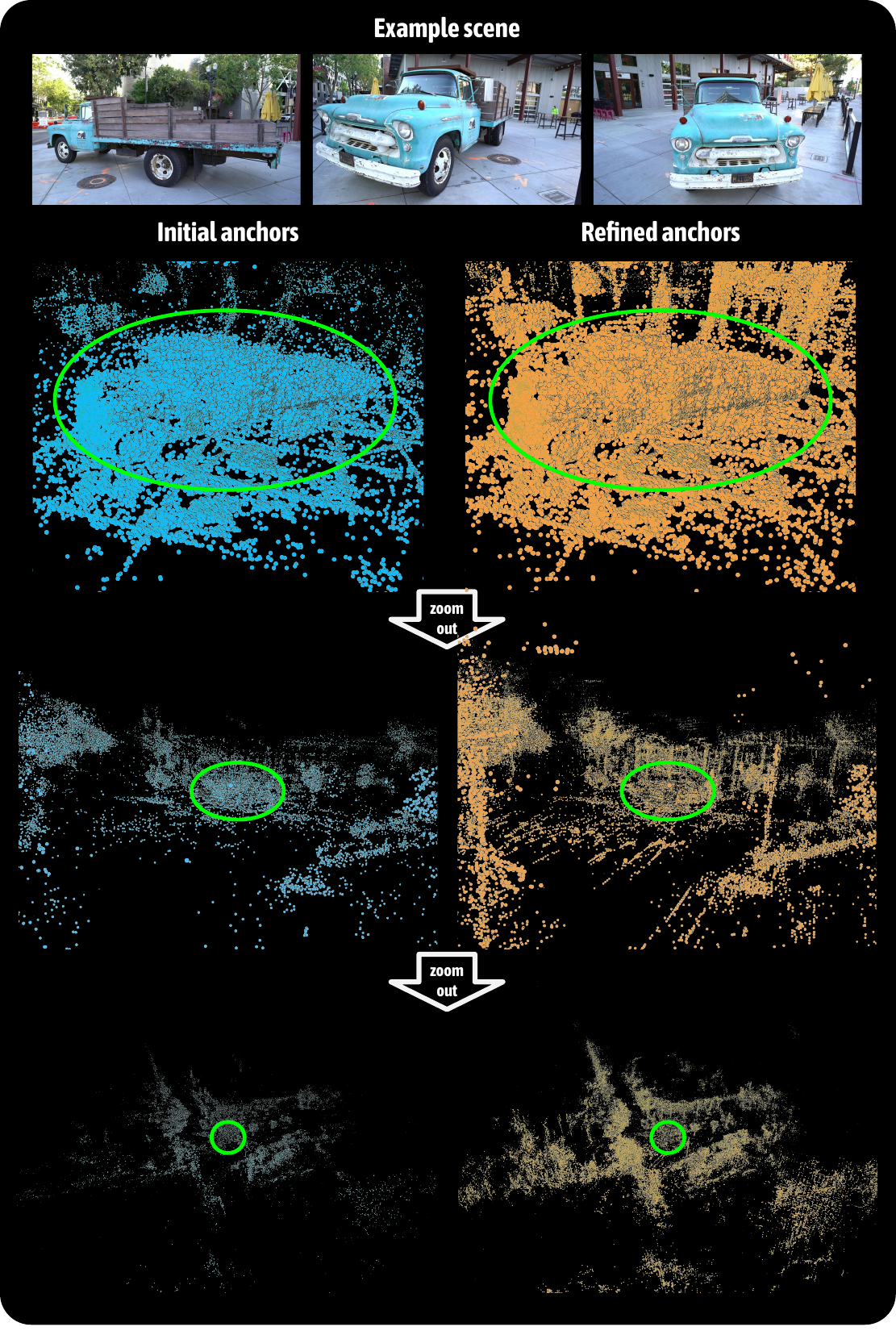}
   \caption{\textbf{Anchor Refinement.} We visualize the \textcolor{cyan}{initial} and \textcolor{orange}{refined} anchor points on the truck scene~\cite{Knapitsch2017}. 
   The truck is highlighted by the \textcolor{green}{circle}.
   Note that the refined points effectively covers surrounding regions and fine-scale structures, leaning to more complete and detailed scene renderings.}
    % The growing and pruning operation play a role in scene completion, which enriches both the structure details and scene completeness. \textcolor{olive}{How to adjust the meshlab screenshot color? \AM{I can do it XD}}}
   
   \label{fig:anchorsgrowing}
\end{figure}

\paragraph{Anchor Refinement.}
% We visualize the effect of anchor refinement process. As shown in Fig.~\ref{fig:anchorsgrowing}, new anchors make up the initial one's shortages in both scene details and large texture-less region.
As briefly discussed in the main paper, the voxelization process suggests that our method may behave sensitive to initial SfM results.
We illustrate the effect of the anchor refinement process in Fig.~\ref{fig:anchorsgrowing}, where new anchors enhance scene details and fill gaps in large texture-less regions and less observed areas.

% \LM{MLP details: activate function;}\\
% \LM{Intern details.}\\
% \LM{CUDA+Depth.}\\

% \LM{Given a bad initial example, such as bicycle.}
\begin{figure*}[]
	\centering
	\includegraphics[width=\linewidth]{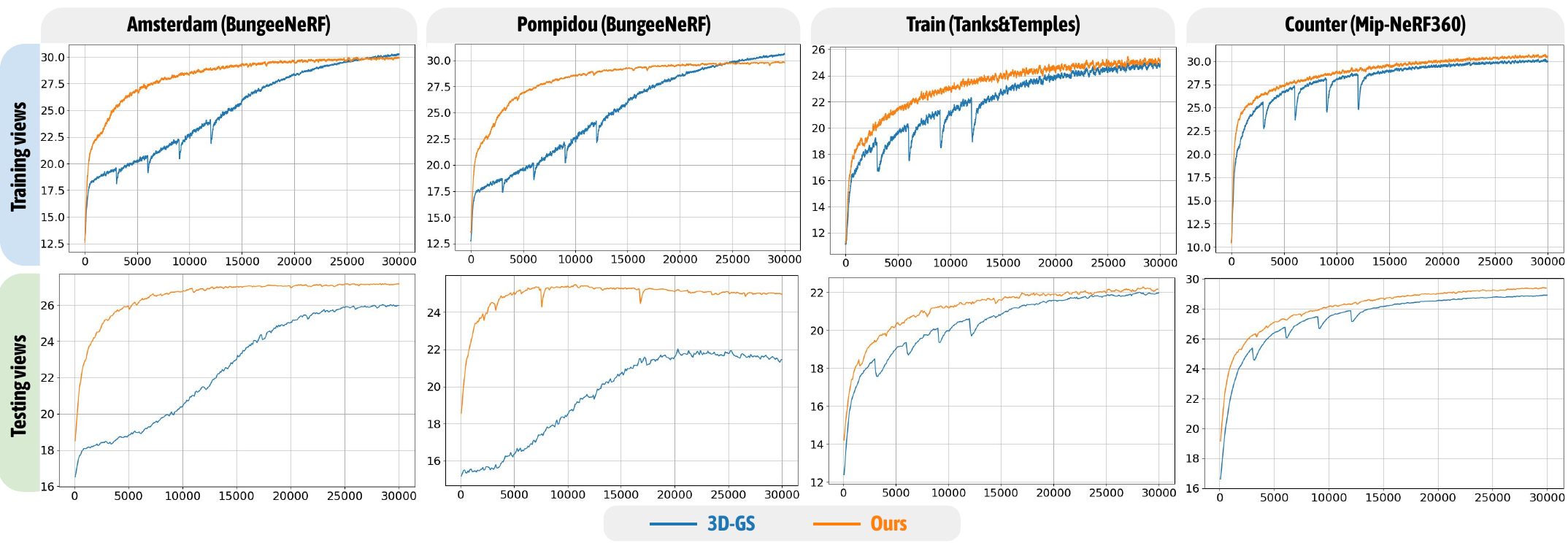}
	\vspace{-20pt}
	\caption{\textbf{PSNR curve of \modelname and 3D-GS~\citep{kerbl3Dgaussians} across diverse datasets~\citep{barron2022mipnerf360,DeepBlending2018,xiangli2022bungeenerf}.} We illustrate the variations in PSNR during the training process for both training and testing views. The orange curve represents \modelname, while the blue curve corresponds to 3D-GS. Our method not only achieves rapid convergence but also exhibits superior performance, marked by a significant rise in training PSNR and consistently higher testing PSNR, in contrast to 3D-GS.  }
	\vspace{-10pt}
	\label{fig:curve}
\end{figure*}
\section{Experiments and Results}

\paragraph{Additional Data Preprocessing.}
% For VR-NeRF and BungeeNeRF datasets, we ran COLMAP~\cite{schoenberger2016sfm} to estimate the camera poses and generate the SfM points for initialization. We use the eye-level subset containing 3 cameras in VR-NeRF dataset for minimum testing. For all other datasets, we follow the original 3D-GS~\cite{kerbl3Dgaussians} and downloaded them from public resources.
We used COLMAP~\cite{schoenberger2016sfm} to estimate camera poses and generate SfM points for VR-NeRF~\cite{VRNeRF} and BungeeNeRF~\cite{xiangli2022bungeenerf} datasets. Both two datasets are challenging in terms of varying levels of details presented in the captures. The VR-NeRF dataset was tested using its eye-level subset with 3 cameras. For all other datasets, we adhered to the original 3D-GS~\cite{kerbl3Dgaussians} method, sourcing them from public resources.

\paragraph{Per-scene Results.}

Here we list the error metrics used in our evaluation in Sec.4 across all considered methods and scenes, as shown in Tab.~\ref{tab:ssim_mip}-\ref{tab:mem_m}.

\begin{table}[]
\centering
\caption{\textbf{SSIM scores for Mip-NeRF360~\cite{barron2022mipnerf360} scenes.} 
% The normalized memory measures how many times the baseline's memory consumption is compared with ours.
}
\vspace{-6pt}
\label{tab:ssim_mip}
\resizebox{\linewidth}{!}{
\begin{tabular}{c|ccccccc}
\toprule
\begin{tabular}{c|c} Method & Scenes \end{tabular}  & bicycle&garden& stump&room&counter&kitchen&bonsai\\
\midrule

\textbf{3D-GS~\cite{kerbl3Dgaussians}}&\textbf{0.771}&\textbf{0.868}&0.775&0.914&0.905&0.922&0.938\\
\textbf{Mip-NeRF360~\cite{barron2022mipnerf360}}&0.685&0.813&0.744&0.913&0.894&0.920&0.941\\
\textbf{iNPG~\cite{muller2022instant}}&0.491&0.649&0.574&0.855&0.798&0.818&0.890\\
\textbf{Plenoxels~\cite{yu2022plenoxels}}&0.496&0.6063&0.523&0.8417&0.759&0.648&0.814\\
\midrule
\textbf{Ours}&0.705&0.842&\textbf{0.784}&\textbf{0.925}&\textbf{0.914}&\textbf{0.928}&\textbf{0.946} \\
\bottomrule

\end{tabular}

}
\end{table}

\begin{table}[]
\centering
\caption{\textbf{PSNR scores for Mip-NeRF360~\cite{barron2022mipnerf360} scenes.} 
% The normalized memory measures how many times the baseline's memory consumption is compared with ours.
}
\vspace{-6pt}
\label{tab:psnr_mip}
\resizebox{\linewidth}{!}{
\begin{tabular}{c|ccccccc}
\toprule
\begin{tabular}{c|c} Method & Scenes \end{tabular}  & bicycle&garden& stump&room&counter&kitchen&bonsai\\
\midrule

\textbf{3D-GS~\cite{kerbl3Dgaussians}}&\textbf{25.25}&\textbf{27.41}&\textbf{26.55}&30.63&28.70&30.32&31.98\\
\textbf{Mip-NeRF360~\cite{barron2022mipnerf360}}&24.37&26.98&26.40&31.63&\textbf{29.55}&\textbf{32.23}&\textbf{33.46}\\
\textbf{iNPG~\cite{muller2022instant}}&22.19&24.60&23.63&29.27&26.44&28.55&30.34\\
\textbf{Plenoxels~\cite{yu2022plenoxels}}&21.91&23.49&20.66&27.59&23.62&23.42&24.67\\
\midrule
\textbf{Ours}& 24.50&27.17&26.27&\textbf{31.93}&29.34&31.30&32.70\\
\bottomrule

\end{tabular}

}
\end{table}

\begin{table}[]
\centering
\caption{\textbf{LPIPS scores for Mip-NeRF360~\cite{barron2022mipnerf360} scenes.} 
% The normalized memory measures how many times the baseline's memory consumption is compared with ours.
}
\vspace{-6pt}
\label{tab:LPIPS_mip}
\resizebox{\linewidth}{!}{
\begin{tabular}{c|ccccccc}
\toprule
\begin{tabular}{c|c} Method & Scenes \end{tabular}  & bicycle&garden& stump&room&counter&kitchen&bonsai\\
\midrule

\textbf{3D-GS~\cite{kerbl3Dgaussians}}&\textbf{0.205}&\textbf{0.103}&\textbf{0.210}&0.220&0.204&0.129&0.205\\
\textbf{Mip-NeRF360~\cite{barron2022mipnerf360}}&0.301&0.170&0.261&0.211&0.204&0.127&\textbf{0.176}\\
\textbf{iNPG~\cite{muller2022instant}}&0.487&0.312&0.450&0.301&0.342&0.254&0.227\\
\textbf{Plenoxels~\cite{yu2022plenoxels}}&0.506&0.3864&0.503&0.4186&0.441&0.447&0.398\\
\midrule
\textbf{Ours}& 0.306&0.146&0.284&\textbf{0.202}&\textbf{0.191}&\textbf{0.126}&0.185\\
\bottomrule

\end{tabular}

}
\end{table}

\begin{table}[]
\centering
\caption{\textbf{Storage size (MB) for Mip-NeRF360~\cite{barron2022mipnerf360} scenes.} 
% The normalized memory measures how many times the baseline's memory consumption is compared with ours.
}
\vspace{-6pt}
\label{tab:mem_mip}
\resizebox{\linewidth}{!}{
\begin{tabular}{c|ccccccc}
\toprule
\begin{tabular}{c|c} Method & Scenes \end{tabular}  & bicycle&garden& stump&room&counter&kitchen&bonsai\\
\midrule

\textbf{3D-GS~\cite{kerbl3Dgaussians}}&1291&1268&1034&327&261&414&281\\
\midrule
\textbf{Ours}& \textbf{248}&\textbf{271}&\textbf{493}&\textbf{133}&\textbf{194}&\textbf{173}&\textbf{258}\\
\bottomrule

\end{tabular}

}
\end{table}

\begin{table}[]
\centering
\caption{\textbf{SSIM scores for Tanks\&Temples~\cite{Knapitsch2017} and Deep Blending~\cite{hedman2018deep} scenes. }}
\vspace{-6pt}

\label{tab:SSIM_two}
	
  \resizebox{\linewidth}{!}{
    \begin{tabular}{c|cc|cc}
    \toprule
\begin{tabular}{c|c} Method & Scenes \end{tabular} & Truck & Train & Dr Johnson & Playroom \\ 
 \midrule
\textbf{3D-GS~\cite{kerbl3Dgaussians}} & 0.879 & 0.802 & 0.899 & \textbf{0.906} \\ 
\textbf{Mip-NeRF360~\cite{barron2022mipnerf360}} & 0.857 & 0.660 &0.901 & 0.900 \\
\textbf{iNPG~\cite{muller2022instant}} & 0.779 & 0.666 & 0.839 & 0.754 \\ 
\textbf{Plenoxels~\cite{yu2022plenoxels}} & 0.774 & 0.663 & 0.787 & 0.802 \\ 
	
\midrule
	 
\textbf{Ours} &\textbf{0.883}&\textbf{0.822}&\textbf{0.907}&0.904\\ 
\bottomrule
\end{tabular}
}
\end{table}

\begin{table}[]
\centering
\caption{\textbf{PSNR scores for Tanks\&Temples~\cite{Knapitsch2017} and Deep Blending~\cite{hedman2018deep} scenes. }}
\label{tab:PSNR_two}
\vspace{-6pt}
	
  \resizebox{\linewidth}{!}{
    \begin{tabular}{c|cc|cc}
    \toprule
\begin{tabular}{c|c} Method & Scenes \end{tabular} & Truck & Train & Dr Johnson & Playroom \\ 
 \midrule
\textbf{3D-GS~\cite{kerbl3Dgaussians}} & 25.19 & 21.10 & 28.77 & 30.04 \\
\textbf{Mip-NeRF360~\cite{barron2022mipnerf360}} & 24.91  & 19.52 & 29.14 & 29.66 \\ 
\textbf{iNPG~\cite{muller2022instant}} & 23.26  & 20.17 & 27.75 & 19.48 \\ 
\textbf{Plenoxels~\cite{yu2022plenoxels}} & 23.22 & 18.93 & 23.14 & 22.98 \\ 
	
\midrule
	 
\textbf{Ours} &\textbf{25.77}&\textbf{22.15}&\textbf{29.80}&\textbf{30.62}\\ 
\bottomrule
\end{tabular}
}
\end{table}

\begin{table}[]
\centering
\caption{\textbf{LPIPS scores for Tanks\&Temples~\cite{Knapitsch2017} and Deep Blending~\cite{hedman2018deep} scenes. }}
\label{tab:LPIPS_two}
\vspace{-6pt}
	
  \resizebox{\linewidth}{!}{
    \begin{tabular}{c|cc|cc}
    \toprule
\begin{tabular}{c|c} Method & Scenes \end{tabular} & Truck & Train & Dr Johnson & Playroom \\ 
 \midrule
\textbf{3D-GS~\cite{kerbl3Dgaussians}} & 0.148 & 0.218 & 0.244 &\textbf{0.241} \\ 
\textbf{Mip-NeRF360~\cite{barron2022mipnerf360}} & 0.159 & 0.354 & \textbf{0.237} & 0.252 \\ 
\textbf{iNPG~\cite{muller2022instant}} &0.274 & 0.386 & 0.381 & 0.465 \\ 
\textbf{Plenoxels~\cite{yu2022plenoxels}} & 0.335 & 0.422 & 0.521 & 0.499 \\ 
	
\midrule
	 
\textbf{Ours} &\textbf{0.147}&\textbf{0.206}&0.250&0.258\\ 
\bottomrule
\end{tabular}
}
\end{table}

\begin{table}[]
\centering
\caption{\textbf{Storage size (MB) for Tanks\&Temples~\cite{Knapitsch2017} and Deep Blending~\cite{hedman2018deep} scenes. }}
\label{tab:mem_two}
\vspace{-6pt}
	
  \resizebox{\linewidth}{!}{
    \begin{tabular}{c|cc|cc}
    \toprule
\begin{tabular}{c|c} Method & Scenes \end{tabular} & Truck & Train & Dr Johnson & Playroom \\ 
 \midrule
\textbf{3D-GS~\cite{kerbl3Dgaussians}} & 578&240&715&515\\ 	
\midrule
	 
\textbf{Ours} &\textbf{107}&\textbf{66}&\textbf{69}&\textbf{63}\\ 
\bottomrule
\end{tabular}
}
\end{table}

\begin{table}[]
\centering
\caption{\textbf{PSNR scores for Synthetic Blender~\cite{mildenhall2021nerf} scenes. }}
\label{tab:PSNR_syn}
\vspace{-6pt}
	
\resizebox{\linewidth}{!}{
    \begin{tabular}{c|cccccccc}
    \toprule
\begin{tabular}{c|c} Method & Scenes \end{tabular} & Mic & Chair & Ship & Materials & Lego & Drums & Ficus & Hotdog \\ 
 \midrule
\textbf{3D-GS~\cite{kerbl3Dgaussians}} & 35.36 & \textbf{35.83} & 30.80 &30.00 &\textbf{35.78} &26.15 & 34.87 & 37.72 \\
\midrule
	 
\textbf{Ours} &\textbf{37.25}&35.28&\textbf{31.17}&\textbf{30.65}&35.69&\textbf{26.44}&\textbf{35.21}&\textbf{37.73}\\ 
\bottomrule
\end{tabular}
}
\end{table}

\begin{table}[]
\centering
\caption{\textbf{Storage size (MB) for Synthetic Blender~\cite{mildenhall2021nerf} scenes. }}
\label{tab:MEM_syn}
\vspace{-6pt}
	
\resizebox{\linewidth}{!}{
    \begin{tabular}{c|cccccccc}
    \toprule
\begin{tabular}{c|c} Method & Scenes \end{tabular} & Mic & Chair & Ship & Materials & Lego & Drums & Ficus & Hotdog \\ 
 \midrule
\textbf{3D-GS~\cite{kerbl3Dgaussians}} & 50&116&63&35&78&93&59&44\\
\midrule
	 
\textbf{Ours} &\textbf{12}&\textbf{13}&\textbf{16}&\textbf{18}&\textbf{13}&\textbf{35}&\textbf{11}&\textbf{8}\\ 
\bottomrule
\end{tabular}
}
\end{table}

\begin{table}[]
\centering
\caption{\textbf{PSNR scores for BungeeNeRF~\cite{xiangli2022bungeenerf} and VR-NeRF~\cite{VRNeRF} scenes. }}
\label{tab:PSNR_m}
\vspace{-6pt}
	
\resizebox{\linewidth}{!}{
    \begin{tabular}{c|cccccc|cc}
    \toprule
\begin{tabular}{c|c} Method & Scenes \end{tabular} & Amsterdam&Bilbao&Pompidou&Quebec&Rome&Hollywood&Apartment&Kitchen\\ 
 \midrule
\textbf{3D-GS~\cite{kerbl3Dgaussians}} & 25.74&26.35&21.20&28.79&23.54&23.25&28.48&29.40\\
\midrule
	 
\textbf{Ours} &\textbf{27.10}&\textbf{27.66}&\textbf{25.34}&\textbf{30.51}&\textbf{26.50}&\textbf{24.97}&\textbf{28.87}&\textbf{29.61}\\ 
\bottomrule
\end{tabular}
}
\end{table}
\begin{table}[]
\centering
\caption{\textbf{Storage size (MB) for BungeeNeRF~\cite{xiangli2022bungeenerf} and VR-NeRF~\cite{VRNeRF} scenes. }}
\label{tab:mem_m}
\vspace{-6pt}
	
\resizebox{\linewidth}{!}{
    \begin{tabular}{c|cccccc|cc}
    \toprule
\begin{tabular}{c|c} Method & Scenes \end{tabular} & Amsterdam&Bilbao&Pompidou&Quebec&Rome&Hollywood&Apartment&Kitchen\\ 
 \midrule
\textbf{3D-GS~\cite{kerbl3Dgaussians}} &1453&1337&2129&1438&1626&1642&202&323 \\
\midrule
	 
\textbf{Ours} &\textbf{243}&\textbf{197}&\textbf{230}&\textbf{166}&\textbf{200}&\textbf{182}&\textbf{48}&\textbf{90}\\ 
\bottomrule
\end{tabular}
}
\end{table}

\paragraph{Training Process Analysis.}
Figure~\ref{fig:curve} illustrates the variations in PSNR during the training process for both training and testing views. Our method demonstrates quicker convergence, enhanced robustness, and better generalization compared to 3D-GS, as evidenced by the rapid increase in training PSNR and higher testing PSNR. Specifically, for the Amsterdam and Pompidou scenes in BungeeNeRF, we trained them with images at \emph{three coarser scales} and evaluated them at a \emph{novel finer scale}. The fact that 3D-GS achieved higher training PSNR but lower testing PSNR indicates its tendency to overfit at training scales.
\newpage
% \subsection{Conv3D for aggregating neighboring anchor features}

% \section{Discussion and Limitations}

% \modelname employs a uniform sparse grid to strategically position neural Gaussians. This approach simplifies the management of Gaussians, including tasks such as querying and clustering. Such an arrangement paves the way for future studies to analyze 3D scenes more efficiently.

% \paragraph{Analogy to proposal network for culving geometry.}

% \paragraph{Scene-specific features in practice.}
% \newpage

%     \small
%     \bibliographystyle{ieeenat_fullname}
%     \bibliography{sup.bib}
% }
\end{document}